\documentclass[sigconf]{acmart}

\begin{document}

\title{Integrating Fuzzy Logic into Deep Symbolic Regression}


\author{Wout Gerdes}
\affiliation{%
  \institution{University of Amsterdam}
  \city{Amsterdam}
  \country{The Netherlands}}
\email{woutgerdes@gmail.com}

\author{Erman Acar}
\affiliation{%
 \institution{IvI \& ILLC, University of Amsterdam}
  \city{Amsterdam}
  \country{The Netherlands}}
\email{erman.acar@uva.nl}


\begin{abstract}
  Credit card fraud detection is a critical concern for financial institutions, intensified by the rise of contactless payment technologies. While deep learning  models offer high accuracy, their lack of explainability poses significant challenges in financial settings. This paper explores the integration of fuzzy logic into Deep Symbolic Regression (DSR) to enhance both performance and explainability in fraud detection. We investigate the effectiveness of different fuzzy logic implications, specifically Łukasiewicz, Gödel, and Product, in handling the complexity and uncertainty of fraud detection datasets. Our analysis suggest  that the Łukasiewicz implication achieves the highest F1-score and overall accuracy, while the Product implication offers a favorable balance between performance and explainability. Despite having a performance lower than state-of-the-art (SOTA) models due to information loss in data transformation, our approach provides novelty and insights into into integrating fuzzy logic into DSR for fraud detection, providing a comprehensive comparison between different implications and methods.
\end{abstract}


\keywords{Deep Symbolic Regression, Fuzzy Logic, Fraud Detection, Explainable AI}

\received{29 June 2024}
\received[revised]{02 October 2024}

\maketitle

\section{Introduction}
\label{sec:introduction}
Credit card fraud poses a significant and growing challenge for financial institutions, amplified by the advent of innovative technologies such as contactless payment \cite{Europol2021SeriousAssessment}. Global losses due to credit card fraud were estimated at \$32.39 billion in 2020 and are projected to exceed \$40 billion by 2027 \cite{S.Nilson2019TheBillion}. The Covid-19 pandemic further accelerated the shift from cash to cashless transactions, intensifying the issue of credit card fraud. To protect customers from fraudulent activities, banks deploy Fraud Detection Systems (FDS) to automatically flag and block suspicious transactions in real-time. Significant advancements in these systems have been achieved through improvements in data quality and the enhanced use and performance of Artificial Intelligence (AI) and Deep Learning (DL) techniques \cite{Cherif2023CreditReview}. Although DL has demonstrated exceptional performance in classification accuracy \cite{Alarfaj2022CreditAlgorithms}, a major limitation is its lack of explainability. \cite{MillOpportunitiesAgenda}. This "black box" nature has hindered the adoption of AI in financial settings, where decisions must be transparent and explainable. Explainable Artificial Intelligence (XAI) offers a potential solution to this problem.

Despite the growing interest in XAI, the intersection of fraud detection and XAI remains underexplored. Recent efforts have taken various approaches to bridge this gap. One approach employs XAI methods to interpret Machine Learning (ML) models post-training using techniques such as SHAP (Shapley Additive Explanations) or LIME (Local Interpretable Model-Agnostic Explanations), which have shown only modest improvements in user trust \cite{JiEXPLAINABLEStudy}. 

Another promising approach involves leveraging Symbolic Regression (SR), which seeks to extract closed-form expressions to describe underlying patterns in the data. These expressions are inherently explainable, resolving transparency issues. To advance SR, Petersen et al. \cite{PetersenDEEPGRADIENTS} combined SR with Reinforcement Learning (RL), resulting in Deep Symbolic Regression (DSR). DSR employs a recurrent neural network (RNN) trained with deep reinforcement learning, where the reward is task-specific, producing expressions tailored to specific problems. These closed-form expressions show high predictive power and transparency, making them a viable solution to many of the previously mentioned issues. DSR utilizes a library of tokens representing features, constants, or mathematical operators to generate expressions. The DSR framework creates a list of tokens subject to constraints, optimizing them using the RNN based on the reward function. An extension to DSR by Visbeek et al. successfully applied DSR to the fraud detection domain, resulting in Deep Symbolic Classification (DSC) \cite{Visbeek2023ExplainableClassification}. DSC adapts DSR for classification tasks and uses the F1-score as the reward metric, offering competitive predictive performance with improved explainability.

In this paper, we propose extensions to the DSR framework by integrating fuzzy logic. Fuzzy logic, based on the fuzzy set theory by Zadeh \cite{Zadeh1965Fuzzy}, categorizes reasoning into multiple levels, similar to human reasoning. Unlike strict classifications, fuzzy logic handles uncertainty and vagueness, making it suitable for real-world complexities. For instance, a person is not simply tall or short but can manifest varying degrees of tallness. Fuzzy logic facilitates smooth transitions between such degrees. In DSC, the output is a closed-form mathematical expression \cite{Visbeek2023ExplainableClassification}. 

Logical implications provide intuitive explanations, since they naturally represent \emph{general rules} e.g., \emph{If transaction amount is high and receiver balance is low then fraud is the case.}

These expressions are more intuitive as they mirror human reasoning. Various formulas derive fuzzy implications from fuzzy sets, and fuzzy logic provides a natural medium for expressing the vagueness within this logical structure. For instance, vague expressions such as amount being high might have a  varying degree of truth  from 0 being false, to 1 being absolutely true. This leads us to the main research question:

\textbf{How can we integrate fuzzy logic into deep symbolic \\regression for fraud detection?} \label{rq:main}

To address our main research question, we elaborate on the following specific sub-research questions:
\begin{enumerate}
    \item What specific fuzzy logic implications are most effective in enhancing the model's ability to handle the inherent complexity and uncertainty in fraud detection datasets? \label{rq:sub1}
    \item What specific choice of fuzzy implication is most effective in DSR? \label{rq:sub2}
    \item Does the fuzzy logic oriented DSR provide intuitive expressions that are easy to interpret? \label{rq:sub3}
    \item What is the trade off between the size of the fuzzy logic formula and performance metrics? \label{rq:sub4}
\end{enumerate}

To assess performance, we compare the proposed framework against current state-of-the-art algorithms as detailed in Table \ref{table:performance_metrics}. Performance evaluation is conducted using accuracy, which is common in real-life fraud detection, and the F1-score, which addresses the inherent imbalance in credit card fraud data. The popular PaySim dataset is utilized \cite{AlonsoLopez-RojasPAYSIM:DETECTION}, offering a controlled environment for comparison. This synthetically generated dataset contains no privacy concerns, as it lacks personally identifiable information (PII). In addition to performance assessment, the explainability of the rule expressions are evaluated. Furthermore, we employ a Pareto front \cite{Pareto} to balance predictive performance and explainability, optimizing the complexity of expressions based on the factors and performance, identifying the most suitable expressions for the task.

\section{Related Work and Background Knowledge}
\label{sec:related_work}
\subsection{Fraud Detection}
Prior research in fraud detection has explored various ML and DL algorithms to address the challenges associated with different types of fraud. Traditional ML approaches, such as Random Forests, k-Nearest Neighbour, Naive Bayes, Logistic Regression, and XGBoost have been extensively studied due to their efficiency in handling structured data \cite{MLinfrauddetection, Hajek2023, Alarfaj2022CreditAlgorithms}. However, these methods often struggle with class imbalance and lack explainability, leading researchers to investigate novel techniques. Recent advancements in DL, particularly convolutional neural networks (CNNs), have shown promise in improving fraud detection by leveraging intricate patterns present in transaction data \cite{Alarfaj2022CreditAlgorithms}. Furthermore, studies have emphasized the importance of considering factors like data imbalance, false positives, and the financial implications of fraud detection systems when designing effective frameworks. Recently, the black box nature of these ML and DL algorithms has been critiqued. The introduction of the Artificial Intelligence Act \cite{AI-act} has formalized the importance of governance in AI applications, highlighting the need for XAI both in general and in fraud detection in particular. 

\subsection{Explainable AI}
 While recent developments in AI promises stellar predictive performance, it often lacks explainability compared to simpler models. XAI aims to increase transparency and explainability of AI systems by providing insights into the decision-making process, model behavior, and factors influencing predictions. This is particularly crucial in sensitive applications such as healthcare, finance, and security, where stakeholders require clear explanations for AI-driven decisions to ensure trust and accountability \cite{MillOpportunitiesAgenda}. These sensitive applications often face regulatory constraints, such as the EU's General Data Protection Regulation (GDPR), requiring enhanced explainability before adopting advanced AI applications \cite{Goodman2016EuropeanExplanation}. XAI can generally be classified into two categories: \textit{potentially interpretable models}, which have some level of inherent transparency (e.g., linear regression, decision trees, k-Nearest Neighbors), and \textit{post-hoc techniques}, which enhance explainability after model training (e.g., Local Interpretable Model-agnostic Explanations (LIME) or SHapley Additive Explanations (SHAP)) \cite{VishnuKuteDeepReview}. 

\subsection{Symbolic Regression}
Symbolic regression (SR) is a method to obtain mathematical expressions that accurately capture information from data. Unlike neural networks, SR aims to provide transparent closed-form expressions, modeling information from the underlying dataset in a form that is more easily interpretable by humans. Given a dataset $(\mathbf{X}, \mathbf{y})$, where each $\mathbf{X}_i$ has inputs $\mathbf{X}_i \in \mathbb{R}^n$ and response $y_i \in \mathbb{R}$, SR attempts to generate a function $f : \mathbb{R}^n \rightarrow \mathbb{R}$ that best fits the dataset, where $f$ is a closed-form mathematical expression \cite{Mundhenk2021SymbolicSeeding}. The formula generated by symbolic classification is much more compact, and tells about the relation between features in contrast to rules generated by decision trees which are often verbose and flat in relation to other features.

Deep symbolic regression (DSR) combines symbolic regression with deep learning techniques, such as RNN's to extract underlying patterns and relationships from data. Unlike traditional symbolic regression, which relies only on predefined mathematical forms, DSR can learn from data and adapt its expressions to capture intricate dependencies, resulting in more accurate and flexible models. The use of an RNN allows for a gradient-based approach to optimize these closed-form expressions, leveraging the performance of certain expressions as the reward for the RNN. With training, the RNN can optimize the reward function per specific mathematical operator, re-evaluating the underlying distribution of each iteration \cite{PetersenDEEPGRADIENTS}. It utilizes mathematical operators from a predefined library \(\lambda\), which includes basic operators (e.g.,\(\div\), \(\cdot\)), constants and features from the dataset. This library enables the generation of closed-form expressions using all available features and performing mathematical operations \cite{PetersenDEEPGRADIENTS, Visbeek2023ExplainableClassification}. Constraints are applied to prevent certain mathematical expressions from being generated, and these constraints can be tailored to improve the DSR framework's performance.
\subsection{Fuzzy Logic}
Fuzzy logic allow for expressing vagueness and uncertainty which can then be incorporated into machine learning \cite{vanKrieken2020AnalyzingOperators}.  By using fuzzy sets, variables are no longer strictly categorized but assigned to multiple categories. This flexibility allows fuzzy logic to generate fuzzy implications in various forms, which are particularly useful in fields like symbolic regression. Different formulas can deduce fuzzy implications from fuzzy sets. In this paper, we focus on Gödel, Product, and Łukasiewicz implications.

Another approach to enable fuzzy logic applications is fuzzy clustering, relying on unsupervised learning instead of domain experts. Fuzzy clustering is the process of acquiring \textit{k} amount of categories, or clusters, per feature of the dataset. Different from human reasoning, fuzzy clustering attempts to gather the categories from the underlying data. Suppose we have a set of categories, \(\{A_1, A_2, \ldots, A_n\}\), and a set of objects \(\{x_1, x_2, \ldots, x_n\}\), and a degree of \(x_i\) belonging to \(A_j\) called \(\mu_{A_j}(x_i)\), where \(0 < \mu_{A_j}(x_i) < 1\). If \(\mu_{A_j}(x_i) = 0\), then \(x_i\) is not a member of \(A_j\), and similarly if \(\mu_{A_j}(x_i) = 1\), then \(x_i\) is a member of \(A_j\) \cite{Askari2020IFDTC4.5:Detection}. Fuzzy clustering is the process of acquiring categories or clusters, \(\{A_1, A_2, \ldots, A_n\}\), per feature of the dataset using clustering algorithms such as \(k\)-means \cite{Alam2009ApplicationDetection}. Each object \(\{x_1, x_2, \ldots, x_n\}\) is assigned to a category, based on its highest membership value, in turn enabling the use of fuzzy logic.

In fuzzy logic, we use strong negation to complement a fuzzy set. If \( a \) is the membership value of an element in a fuzzy set, the strong negation is defined as $N(a) = 1 - a$

T-norms (triangular norms) model the conjunction (AND) operation in fuzzy logic and generalize the intersection operation in classical set theory.

T-conorms (triangular conorms) model the disjunction (OR) operation in fuzzy logic and generalize the union operation in classical set theory \cite{hajek2013metamathematics}.

In fuzzy logic, implications play a crucial role in reasoning. There are two primary types of implications  \cite{hajek2013metamathematics}:
\begin{itemize}
    \item \textbf{S-implications} are derived from the residuation principle and are defined using T-norms.
    \item \textbf{R-implications} are defined using a residual function and T-conorms.
\end{itemize}
For the Gödel, Product, and Łukasiewicz implications, the specific formulas can be found in Table \ref{table:implications}.

\begin{table*}[h!]
\centering
\begin{tabular}{|c|c|c|c|}
\hline
\textbf{Operation Type} & \textbf{Gödel} & \textbf{Product} & \textbf{Łukasiewicz} \\
\hline
\textbf{T-norm} & \( T_{\text{gd}}(a, b) = \min(a, b) \) & \( T_{\text{pr}}(a, b) = a \cdot b \) & \( T_{\text{lk}}(a, b) = \max(a + b - 1, 0) \) \\
\hline
\textbf{T-conorm} & \( S_{\text{gd}}(a, b) = \max(a, b) \) & \( S_{\text{pr}}(a, b) = a + b - a \cdot b \) & \( S_{\text{lk}}(a, b) = \min(a + b, 1) \) \\
\hline
\textbf{S-implication} & 
\( I_{\text{gd}}(a, c) = \max(1 - a, c) \) & 
\( I_{\text{pr}}(a, c) = 1 - a + a \cdot c \) & 
\( I_{\text{lk}}(a, c) = \min(1 - a + c, 1) \) \\
\hline
\textbf{R-implication} & 
\( I_{\text{gd}}(a, c) = \begin{cases} 
1 & \text{if } a \leq c \\ 
c & \text{otherwise} 
\end{cases} \) & 
\( I_{\text{pr}}(a, c) = \begin{cases} 
1 & \text{if } a \leq c \\ 
\frac{c}{a} & \text{otherwise} 
\end{cases} \) & 
\( I_{\text{lk}}(a, c) = \min(1 - a + c, 1) \) \\
\hline
\end{tabular}
\caption{T-norms, T-conorms, S-implications, and R-implications for Gödel, Product, and Łukasiewicz \cite{vanKrieken2020AnalyzingOperators}.}
\label{table:implications}
\end{table*}

These fuzzy implications are required when working with fuzzy sets to capture underlying patterns from the data.

\subsection{Integrating Fuzzy Logic in Symbolic Regression}
Symbolic regression traditionally focuses on generating mathematical expressions that best fit the data. However, using logical operators instead of purely mathematical ones can improve explainability and enhance alignment with human reasoning. 
The language of logic is more natural for explaining phenomena, as logical operators like "and", "or", and "not" are closer to daily language. For instance, rules in the form of implications can be easily understood as "if-then" statements, which are common in human reasoning. Historically, logic stems from natural language, originating with syllogisms by Aristotle, and abstracting important aspects of human reasoning \cite{SyllogismAristotle}. The Łukasiewicz implications and logic are essentially a modern reconstruction of these syllogisms \cite{SyllogisticLukas}. This historical context indicates the similarities of logic with how humans understand and describe the world \cite{Russell1919-RUSITM}. Therefore, fuzzy logic, which incorporates similar logical operators, offers a more intuitive and explainable framework compared to purely mathematical expressions. Fuzzy logic is particularly suitable for application in symbolic regression because it retains the numerical foundation of symbolic regression while leveraging the explainability of logical expressions. By incorporating fuzzy sets and implications, we can handle vagueness and uncertainty more effectively while aligning the explanation with human reasoning.

\section{Methodology}
\label{sec:methodology}

\subsection{Model}
We adapt DSR to integrate fuzzy logic, extending the existing DSR approach by integrating fuzzy implications into the expression generation and evaluation stages.
Traditionally, the DSR framework utilizes a generated traversal of tokens to create a binary syntax tree, which is then used to generate expressions of these tokens and test their performance on the dataset. Our approach augments this framework to integrate fuzzy logic implications \cite{PetersenDEEPGRADIENTS}.

To facilitate the use of fuzzy logic in our model, we transform relevant features into fuzzy sets, capturing  degrees of membership such as "very large" or "somewhat small." This transformation allows the model to better manage the uncertainty and variability present in financial transactions. Fuzzy logic formulas, as shown in Table \ref{table:implications}, are employed in generating the expressions. 

We test and compare three different fuzzy logic implications: Gödel, Product, and Łukasiewicz. Fuzzy membership functions determine the degree to which a transaction belongs to each class, providing a spectrum of likelihood rather than a binary decision. We implement and evaluate the S-implication, together with the T-norm, T-conorm, and strong negator for these three fuzzy implications. Note that the S-implications, in contrast to R-implications, are the ones that generalize material implications, since they use the T-conorm in their construction \cite{PetersenDEEPGRADIENTS}. Therefore, we utilize the S-implication and disregard the R-implication. Additionally, we compare a combination of all three fuzzy logic implications, resulting in a comparison between four function sets.

Our implementation is flexible, allowing the use of any\\ S-implications, T-norms, T-conorms, and strong negators. To research the impact of implementing the S-implication as the root of the expression, we modify the DSR package. We add a constraint ensuring the S-implication is always the parent of a T-norm, T-conorm, or strong negator. This ensures the S-implication is never a child of any other function. However, due to the nature of syntax trees generated from functions and features, not every traversal includes the S-implication. To address this, the token for the S-implication was added to the traversal at index 0 if it was not yet included. If the traversal already contains the S-implication token, we swap its position with the token at index 0.

Through reinforcement learning techniques, we train the model to optimize for F1-score, following the research of \cite{Visbeek2023ExplainableClassification}. During training, a RNN produces mathematical expressions and evaluates them based on their ability to describe the dataset, a measure known as "fitness". This fitness score links to a reward, which trains the RNN through a risk-seeking policy gradient algorithm. The risk-seeking policy gradient algorithm is defined in Equation \ref{eq:risk-seeking}. This risk-seeking policy is utilized to improve the model's functioning. The risk-seeking policy focuses on improving the reward of the top \(\epsilon\) fraction of samples, disregarding samples below the threshold. This method aims to enhance the highest performing expressions, even if it results in a lower average performance \cite{PetersenDEEPGRADIENTS}. We use the F1-score as the main reward function, although we also test the F2-score as the reward function. The policy gradient algorithm adjusts the probabilities of sampling an expression according to its corresponding reward. Specifically, expressions with higher fitness scores receive higher rewards, which in turn increase their likelihood of being sampled in subsequent iterations. This method ensures that the RNN progressively favors expressions that better fit the data.

\begin{equation}
J_{risk}(\theta ; \epsilon) \equiv \mathbb{E}_{\tau \sim p(\tau \mid \theta)} [R(\tau) \mid R(\tau) \geq R_{\epsilon}(\theta)]
\label{eq:risk-seeking}
\end{equation}

By integrating fuzzy logic into DSR, our approach aims to improve the robustness and explainability of fraud detection systems.

\subsection{Data}
In this research, we utilize the popular and well-researched synthetically generated dataset, PaySim \cite{AlonsoLopez-RojasPAYSIM:DETECTION}. This dataset consists of simulated transactions based on a distribution of proprietary real transactions. Due to inherent privacy concerns with real-life datasets, this is the most suitable alternative available. PaySim contains over 6 million financial transactions over a period of one month, with a fraud rate of 0.13\%. Given the true/false rate of 0.13/99.87, this is a highly imbalanced dataset and should be handled accordingly.

The dataset includes the following features \cite{Visbeek2023ExplainableClassification}:
\begin{itemize}
    \item \textbf{step} - unit of time; one step corresponds to one hour of time,
    \item \textbf{type} - a categorical feature with values: cash-in, cash-out, debit, payment, or transfer,
    \item \textbf{amount} - amount of the transaction,
    \item \textbf{nameOrig} - name of the customer,
    \item \textbf{oldbalanceOrg} - customer’s balance before the transaction,
    \item \textbf{newbalanceOrig} - customer’s balance after the transaction,
    \item \textbf{nameDest} - name of the recipient,
    \item \textbf{oldbalanceDest} - recipient’s balance before the transaction,
    \item \textbf{newbalanceDest} - recipient’s balance after the transaction,
    \item \textbf{isFlaggedFraud} - an indicator of whether the transaction has been flagged as fraudulent in the simulation,
    \item \textbf{isFraud} - an indicator of the transaction being legitimate or fraudulent.
\end{itemize}

The isFraud feature represents the target variable, which serves as the ground truth for training our model.

Per the creators' request \cite{AlonsoLopez-RojasPAYSIM:DETECTION}, the features oldbalanceOrig, newbalanceOrig, oldbalanceDest, newbalanceDest should not be used directly. This is because fraudulent transactions are cancelled, meaning the balances of the originator and recipient remain unchanged. Using these features directly would result in artificially high accuracy, failing to accurately classify real fraud detection cases.

Instead, these features provide insights into typical transaction behavior. To avoid direct usage, we use imputation techniques to generate new features that capture balance changes without relying on the original balance columns. Specifically, we engineer features such as newbalanceDest, oldbalanceOrig based on the transaction amount plus oldbalanceDest, newbalanceOrig. Hourly, daily, and monthly features are also derived from the step feature. The feature is\_workday is created based on the transaction volume distribution over a week, assuming the five highest volume days correspond to weekdays and the lowest two to weekends.

Categorical features are one-hot encoded, converting textual categorical data into binary columns, making them suitable for DSR. Although these features are not fuzzy sets, they are still useable for this research.

To improve model robustness and generalizability, we introduce random noise into the dataset. For each feature column, Gaussian noise is added using a method where the standard deviation of the target column is multiplied by a noise level parameter of 0,05. This noise simulates real-world data imperfections, preventing the model from overfitting to clean training data. This approach enhances the model's resilience and its ability to generalize to new, unseen data.

We also incorporate transaction history by creating rolling average and rolling max features for the amount column. Using 3 and 7 transaction windows, we calculate the mean and max transaction amounts for each recipient (nameDest), helping to identify discrepancies based on transaction history.

To transform the dataset into fuzzy sets, we divide the dataset into percentiles. Each feature is represented as a fuzzy set according to these percentiles. We calculate the 20th, 40th, 60th, and 80th percentiles for each feature, segmenting the data into five ranges. For each feature value, we assign a fuzzy membership value based on its percentile. For example, values less than or equal to the 20th percentile receive a membership value of 0.2, the 20th to 40th percentile values receive 0.4, and so on. This transformation is applied to all relevant features, ensuring they are represented as fuzzy sets, facilitating the integration of fuzzy logic into our model \cite{Razooqi2016CreditNetwork}.  The features used are all either transformed into fuzzy sets, or binary values. 

\subsection{Hyperparameter Tuning}
To prepare the dataset for machine learning tasks, it is necessary to partition it into distinct sets for training, validation, and final testing. This process ensures that the model's performance can be accurately assessed and optimized. Due to computational constraints and efficiency considerations, hyperparameter tuning is performed on a subset of the data, while the final testing is conducted on another separate subset.

Initially, a sample of one million transactions is extracted from the dataset. It is essential to ensure that the balance of fraudulent and non-fraudulent transactions in this sample accurately reflects that of the full dataset. This maintains the representativeness of the sample and ensures that any insights or models derived from it remain applicable to the entire dataset. For hyperparameter tuning, we partition the sample into training and validation sets using a 70/30 split ratio, respectively. We perform this split in a stratified manner, preserving the balance of fraudulent and non-fraudulent transactions across the sets. Stratification helps mitigate the risk of introducing biases during hyperparameter tuning, ensuring a comprehensive evaluation of the performance across different classes.

A common concern when utilizing DSR is its sensitivity to hyperparameters \cite{PetersenDEEPGRADIENTS}. Although computational constraints limited extensive testing and optimization of hyperparameters, key parameters were tuned. Important hyperparameters tested are:

\begin{itemize}
    \item \textbf{Sigmoid threshold} - defines the cutoff value for the sigmoid activation function, squashing the output into a range between 0 and 1.
    \item \textbf{Learning rate} - controls the step size at each iteration while moving towards a minimum of the loss function.
    \item \textbf{Entropy weight} - scales the entropy term in the loss function, promoting exploration by encouraging a more uniform probability distribution.
    \item \textbf{Risk factor} (\(\epsilon\)) - used by the risk-seeking policy gradient.
    \item \textbf{Batch size} - number of training samples used in one iteration of model training.
    \item \textbf{\textit{N} samples} - the total number of samples to generate.
\end{itemize}

The higher the value for \textit{N} samples, the more batches DSR uses to improve the found expressions. For each run, a sample size of 200,000 was used. This ensures minimal progress in the last quarter of iterations, balancing computational feasibility and optimal performance.

We perform hyperparameter tuning on the one million transaction subset. Each hyperparameter setting was tested on 16 runs, seeded from seed 0 to seed 15 to ensure comparability. The average F1 score and highest F1 score were compared across all hyperparameter settings. The best performing hyperparameter configuration was chosen for the model. 

For the final evaluation of the model's performance, a separate sample comprising two million transactions was extracted from the dataset. Similar to the initial sample, the balance of fraudulent and non-fraudulent transactions was maintained. The two-million transaction subset is split into training and test sets using a ratio of 70/30. This split was also stratified based on the fraud column, ensuring proportional representation of fraudulent and non-fraudulent transactions in both sets.

To eliminate potential biases introduced by the ordering of the dataset, both the subsampling and splitting processes were accompanied by shuffling. Shuffling ensures that each subset is a random representation of the original dataset, preventing any unintentional patterns or correlations from influencing the model's performance. Additionally, to ensure reproducibility, both the shuffling and splitting procedures were performed using a specific seed value, allowing for consistent results across multiple runs and ensuring the reproducibility of the experimental setup.

\subsection{Evaluation}
Evaluation of the proposed framework can be approached from two perspectives: classification performance and explainability performance. Classification performance represents how accurately the framework detects fraudulent transactions. To correctly assess this performance, we use two different measures: accuracy and F1 score.
Accuracy is a standard gauge for comparing the proposed framework to current industry standards. However, due to the highly imbalanced nature of our dataset, accuracy can be misleading. A high accuracy is easily achieved by marking all transactions as non-fraudulent, making it less meaningful in our context. Therefore, the F1 score is also used to address this imbalance problem \cite{Jeni2013FacingMetrics}.\\

\noindent
\begin{minipage}{0.22\textwidth}
    \begin{equation}
    \text{F1-score } = \frac{2 \cdot P \cdot R}{P + R}
    \label{eq:f1_score}
    \end{equation}
\end{minipage}\hfill
\begin{minipage}{0.22\textwidth}
    \begin{equation}
    \text{F2-score } = \frac{5 \cdot P \cdot R}{4 \cdot P + R}
    \label{eq:f2_score}
    \end{equation}
\end{minipage}\\

The F1 score is the harmonic mean of precision and recall. It balances the importance of both metrics, which is crucial for our imbalanced dataset. The RNN will also be optimized for the F1 score. In highly imbalanced datasets, recall is often more important than precision. Hence, the F2 score \cite{Jeni2013FacingMetrics} is also investigated, where recall is twice as important as precision.

Explainability of the framework is assessed using a Pareto front. A Pareto front evaluates the trade-off between expression complexity and predictive performance. Operators in the DSR framework are weighted based on their pre-determined complexity. For instance, the complexity of T-norm, T-conorm, and strong negator are set to one, while S-implication is set to two. The complexity of a generated expression is the sum of its components' complexities. The Pareto front represents the set of solutions where no other expression achieves superior performance for the corresponding complexity \cite{Smits2005}. This helps determine a balance between performance and explainability, resulting in a final expression that is both effective and explainable.

Comparing performance is challenging due to the novelty of the proposed method, and the reduction in information in the features by transforming them into fuzzy sets. Previous research on this dataset uses k-NN, SVM, RF, and XGBoost \cite{Hajek2023}, while \cite{Visbeek2023ExplainableClassification} used DSC. These methods are used for rough comparison to evaluate if integrating fuzzy logic into the DSR framework might be useful.
\begin{table}[htbp]
    \centering
    \caption{SOTA performance metrics}
    \label{table:performance_metrics}
    \begin{tabular}{lcccc}
        \hline
        \textbf{Method} & \textbf{F1} & \textbf{Accuracy} & \textbf{Precision} & \textbf{Recall} \\
        \hline
        k-NN & 0.16 & 0.99 & 0.087 & 0.87 \\
        SVM & 0.47 & 0.99 & 0.95 & 0.31 \\
        RF & 0.84 & 0.99 & 0.92 & 0.78 \\
        XGBoost & 0.84 & 0.99 & 0.88 & 0.81 \\
        DSC & 0.78 & 0.99 & 0.95 & 0.67 \\
        \hline
    \end{tabular}
    \vspace{0.5em}
    \caption*{Sources: k-NN, SVM, RF, XGBoost \cite{Hajek2023}, DSC \cite{Visbeek2023ExplainableClassification}}
\end{table}
\section{Results}
\subsection{Performance of Different Implications}
The results compare the performance of the Gödel, Product, and Łukasiewicz implications in the DSR framework. Table \ref{table:bestresults} shows the highest score achieved by any expression found across 16 seeds, and Table \ref{table:avgresults} presents the average performance of the best-performing expressions per seed. Each seed utilizes a different slice of the sample, resulting in different formulas and expressions. All expressions are evaluated on the test set.

\begin{table}[!htbp]
\caption{Performance metrics of the best single expression per method. Bold indicates the best score in each column.}
\begin{tabular}{lcccc}
\toprule
\textbf{Method} & \textbf{F1-score} & \textbf{Accuracy} & \textbf{Precision} & \textbf{Recall} \\
\midrule
Product & 0.163 & \textbf{0.998} & \textbf{0.161} & 0.165 \\
Łukasiewicz & \textbf{0.193} & 0.997 & 0.154 & \textbf{0.256} \\
Gödel & 0.073 & 0.996 & 0.050 & 0.131 \\
Combined & 0.177 & 0.996 & 0.126 & 0.295 \\
\bottomrule
\end{tabular}
\label{table:bestresults}
\end{table}
Table \ref{table:bestresults} highlights the best single expression's performance per method. The Łukasiewicz implication achieves the highest F1-score and recall, while the Product implication has the highest precision and accuracy.
\begin{table}[!htbp]
\caption{Average performance of the best expression per seed. Bold indicates the best score in each column.}
\begin{tabular}{lcccc}
\toprule
\textbf{Method} & \textbf{F1-score} & \textbf{Accuracy} & \textbf{Precision} & \textbf{Recall} \\
\midrule
Product & 0.157 & 0.997 & 0.183 & 0.141 \\
Łukasiewicz & \textbf{0.169} & \textbf{0.998} & \textbf{0.194} & 0.164 \\
Gödel & 0.072 & 0.996 & 0.131 & 0.050 \\
Combined & 0.158 & 0.998 & 0.152 & \textbf{0.180} \\
\bottomrule
\end{tabular}
\label{table:avgresults}
\end{table}
Table \ref{table:avgresults} presents the average performance of the best expressions per seed. The Łukasiewicz implication leads with the highest average F1-score and accuracy, indicating robust performance across both the best expression and average expression.

Currently, the best performing expression is Equation \ref{eq:lukasiewicz_formula}, with performance results in Table \ref{table:bestresults}. The best-performing fuzzy logic expression uses the Łukasiewicz implication. It involves nested implications and norms, leveraging features indicating balance of the receiving address both before and after the transaction, the type of transaction, and the maximum transaction amount of the past 7 transactions to predict fraud. Abbreviations are used to improve readability of the formula. \textit{NBD} corresponds to newbalanceDest, \textit{OBD} is oldbalanceDest, \textit{MD7} is maxDest7 and \textit{C\_O} is type\_CASH\_OUT.  For formal details of fuzzy logic, we refer to  \cite{hajek2013metamathematics}.

\begin{equation}
\begin{aligned}
\forall x, y \quad \neg \Bigg( 
    & \bigg( 
        \Big( 
            \Big( 
                \neg \Big( \neg \big( \text{NBD} \big) \Big) \otimes \big( \text{NBD} \big) 
            \Big) \otimes
    \Big( 
        \Big( 
            \text{OBD} \Big) \\
            &\rightarrow \Big( \neg \Big( 
                \Big( 
                    \text{OBD} \Big) \rightarrow 
                    \big( \text{MD7} \big) 
                \Big) 
            \Big) 
        \Big)
    \otimes 
    \big( \text{C\_O} \big) 
    \bigg) \rightarrow 
    \big( \text{OBD} \big)
\Bigg)
\end{aligned}
\label{eq:lukasiewicz_formula}
\end{equation}

The new balance of the destination account (NBD) compared with its old balance (OBD) can indicate anomalies. For example, if the new balance significantly exceeds typical values following a series of large transactions, it could be suspicious.

The nested implications involving the old balance of an account (OBD) and the maximum of the past 7 transactions (MD7) evaluate how recent large transactions relate to the current balance. If the OBD does not logically align with past transaction patterns, it might signal fraud.

\subsection{F2 Score}
We utilize the Łukasiewicz implication for further research into the usage of F1 versus F2 scores as the reward function, as shown in Table \ref{table:f1f2score}. We train the model using F2 as a reward score and compare it against the model trained using the F1 as the reward score, testing whether prioritizing recall over precision improves performance. 
\begin{table}[!htbp]
\centering
\caption{Comparison of F1 versus F2 score.}
\begin{tabular}{lcccc}
\toprule
\textbf{Score Type} & \textbf{F1-score} & \textbf{Accuracy} & \textbf{Precision} & \textbf{Recall} \\
\midrule
F1 Score (best) & 0.193 & 0.997 & 0.154 & 0.256 \\
F2 Score (best) & 0.135 & 0.998 & 0.150 & 0.123 \\
F1 Score (avg) & 0.169 & 0.998 & 0.194 & 0.164 \\
F2 Score (avg) & 0.126 & 0.991 & 0.493 & 0.082 \\
\bottomrule
\end{tabular}
\label{table:f1f2score}
\end{table}
In Table \ref{table:f1f2score}, we compare models trained with F1 and F2 scores as the reward function. Prioritizing recall (F2 score) slightly improves accuracy but decreases the overall F1-score, indicating a trade-off between precision and recall. Utilising F2 score results in a high average precision, which is contrary to expectations and will be discussed further in Section \ref{sec:discussion}.

\subsection{Constrained Implementation}
We also test the Łukasiewicz implication when it is constrained as the root node and compare this against the unconstrained implementation, as shown in Table \ref{table:constraint}. The model is trained while being constrained such that the S-implication occurs in the root node of the expression, and compare this against the unconstrained implementation. Abbreviations are used to improve readability, where Unc corresponds to the unconstrained implementation, and Con corresponds to the constrained implementation.

\begin{table}[!htbp]
\centering
\caption{Performance comparison of unconstrained\\ versus constrained implementation.}
\begin{tabular}{lcccc}
\toprule
\textbf{Constraint} & \textbf{F1-score} & \textbf{Accuracy} & \textbf{Precision} & \textbf{Recall} \\
\midrule
Unc (best) & 0.193 & 0.997 & 0.154 & 0.256 \\
Con (best) & 0.163 & 0.998 & 0.161 & 0.165 \\
Unc (avg) & 0.169 & 0.998 & 0.194 & 0.164 \\
Con (avg) & 0.133 & 0.996 & 0.195 & 0.138 \\
\bottomrule
\end{tabular}
\label{table:constraint}
\end{table}

Table \ref{table:constraint} highlights that the unconstrained implementation generally outperforms the constrained ones, suggesting that allowing more flexibility in the structure  can lead to better performance. 



\section{Discussion}
\label{sec:discussion}

\noindent\textbf{Performance Comparison.} The performance analysis indicates that the Łukasiewicz and Product fuzzy logic implications perform comparatively well, with the Łukasiewicz implication achieving the highest F1-score and overall accuracy. In contrast, the Gödel implication significantly lags behind, highlighting the importance of selecting the appropriate fuzzy logic implication for fraud detection tasks. The combination of different fuzzy logic implications does not yield performance improvements, suggesting that the unique characteristics of each fuzzy logic implication are best utilized independently. This finding underscores further exploration of individual fuzzy logic implications to enhance their performance.

\noindent\textbf{Performance against State-Of-The-Art.}
The performance of all fuzzy logic implications within the DSR framework is lower than SOTA methods such as k-NN, SVM, RF, and XGBoost, as detailed in Table \ref{table:performance_metrics}. It is important to note that although these methods are trained on the same dataset, different transformations are used. In this research, all data is compressed into values between [0, 1]. In the research regarding other methods, the data is kept in its original format, ensuring more information is retained in the dataset. Therefore, a comparison between these performance metrics should be considered more of a proof of concept for integrating fuzzy logic into the DSR framework then an accurate performance comparison. Furthermore, this outcome was anticipated due to the transformation of transaction data into fuzzy sets, which results in substantial information loss. While the current approach compresses data into five categories based on percentiles, this simplification inherently limits the information density of the original data. Enhancements in the way fuzzy sets are defined, such as through unsupervised fuzzy clustering or using transformations like logarithmic scaling, could help retain more information and improve performance. 

Analyzing the best-performing expression from DSC \cite{Visbeek2023ExplainableClassification}, we see that features like \textit{maxDest7} and \textit{amount} capture information about transaction values, which fuzzy sets currently fail to represent adequately. For instance, the \textit{maxDest7} feature is subtracted from \textit{amount} to highlight anomalies in the transaction history. This nuanced relationship is lost when data is compressed into fuzzy sets. By utilizing fuzzy clustering, we might better capture the intricate balance between current and previous transactions, potentially enhancing the model's performance.

\noindent\textbf{F1 versus F2 Score.}
Training the DSR framework with the F2 score as the reward function demonstrates the impact of emphasizing recall over precision. In fraud detection, it is crucial to identify as many fraudulent transactions as possible, even if it means a higher false-positive rate. The F2 score, which weighs recall twice as much as precision, aligns with this priority. As shown in Table \ref{table:f1f2score}, prioritizing recall (F2 score) slightly improves accuracy for the best performing expression, but decreases the overall F1-score. On average, utilising the F2 score as the reward function yields worse performance compared to the F1 score.

Interestingly, utilizing the F2 score results in a very high average precision, contrary to expectations. The expectation when utilizing the F2 score was to achieve higher recall at the expense of precision. This is because the F2 score places more emphasis on recall, encouraging the model to identify as many fraudulent transactions as possible. Typically, this approach would result in a larger number of false positives, thereby reducing precision. Due to the nature of DSR, expressions can be found that achieve a high F1 score, but are not in line with the expectations.

The best performing expression extracted using the F2 score as a reward function can be seen in Expression \ref{eq:F2_formula}. OBD stands for oldBalanceDest, TF stands for type\_TRANSFER and MD7 represents maxDest7. The T-norm, T-conorm and S-implication all correspond to the Łukasiewicz implications.

\begin{equation}
\begin{split}
\neg \Big( \oplus \Big(\text{OBD}, \rightarrow \Big( \text{TF}, \text{MD7}\Big)\Big)\Big)
\end{split}
\label{eq:F2_formula}
\end{equation}

\begin{figure}[!htbp]
    \centering
    \includegraphics[width=1\linewidth]{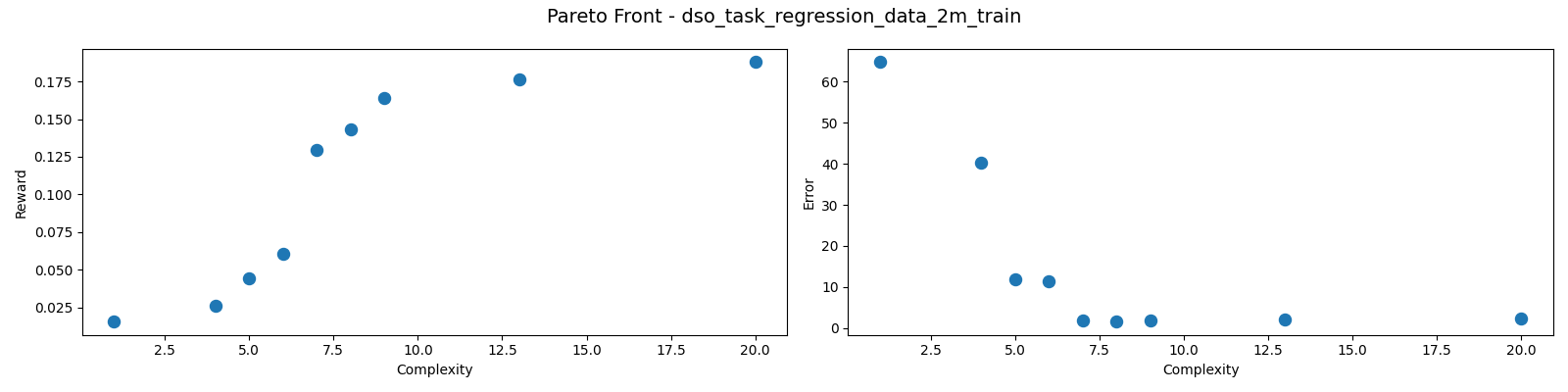}
    \caption{Pareto front showcasing the complexity and performance curve for the Łukasiewicz implication}
    \label{figure:pareto}
\end{figure}

\noindent\textbf{Constrained Implementation.}
Constraining the implementation of fuzzy implications results in worse performance compared to the unconstrained version. This finding suggests that the flexibility to explore a wide range of expression structures is more suitable for optimal performance. By constraining the model to use specific logical rules, such as placing the S-implication at the root node, the model's ability to discover effective patterns is limited. This insight challenges the hypothesis that strict logical constraints always lead to better performance, highlighting the importance of allowing the model to explore diverse expression configurations. However, the expression still generates interesting insights into the implementation of fuzzy logic into the DSR package. Expression \ref{eq:constrained} shows the best performing expression generated with the given constraints. OBD stands for oldBalanceDest, NBD stands for newBalanceDest and  TF stands for type\_TRANSFER. The T-norm, T-conorm and S-implication all correspond to the Łukasiewicz implications.
\begin{equation}
\begin{split}
\rightarrow \Big( \oplus \Big(\text{TF}, \neg \Big(\text{TF}\Big)\Big), \otimes \Big(\otimes\Big(\neg\Big(\neg\Big(\text{NBD}\Big)\Big),\\
\otimes \Big( \neg \Big( \text{OBD}, \neg \big( \text{OBD} \Big) \Big) \Big),\text{NBD} \Big) \Big)
\end{split}
\label{eq:constrained}
\end{equation}
\noindent\textbf{Explainability and Trade Off.}
Fuzzy logic offers intuitive rule-based expressions that are easier to understand compared to traditional black-box models. However, there is a trade-off between the complexity of these expressions and their predictive performance. More complex expressions can capture nuanced patterns in the data but may become less explainable. The Pareto front, shown in Figure \ref{figure:pareto}, helps visualize this trade-off, allowing for the selection of expressions that balance performance and complexity. This balance is crucial for ensuring that the fraud detection system is both effective and explainable.

The best-performing formula using the Łukasiewicz implication is shown in Expression \ref{eq:lukasiewicz_formula}. This formula leverages several fuzzy logic operations to evaluate the likelihood of a transaction being fraudulent. The operations involve taking the minimum and maximum of various terms, which correspond to different transaction features such as \textit{newBalanceDest}, \textit{oldbalanceDest}, and \textit{maxDest7}, combined with transaction type \textit{type\_CASH\_OUT}. The formula's structure has a series of logical conditions and implications that aim to capture the complexities of fraud detection. The complexity of this formula makes it difficult to interpret. The nested operations and multiple levels of logical implications obscure the direct relationships between input features and the final output. Understanding how each part contributes to the overall decision requires analysing each nested operation, which is not improving explainability.

In contrast, the best-performing expression using the Product implication is simpler and more explainable. We use abbreviations for convenience; NBD is newBalanceDest, OBD is oldBalanceDest. 
\begin{equation}
\begin{split}
\otimes \Big( \text{NBD}, \otimes \Big( \neg \Big(\oplus \Big(\text{OBD}, \text{OBD}\Big)\Big), \oplus \Big(\text{OBD}, \text{OBD}\Big)\Big)
\end{split}
\label{eq:product_implication}
\end{equation}
The inclusion of the new balance of the destination (NBD) directly in the conjunction signifies its importance in determining the likelihood of fraud. A significant change in the new balance of the destination account could indicate unusual activity, especially if it does not align with typical transaction patterns.

The old balance (OBD) appears in both the negation and the disjunction operations, suggesting that the old balance of the account is crucial for understanding its normal behavior.

\begin{figure}
    \centering
    \includegraphics[width=1\linewidth]{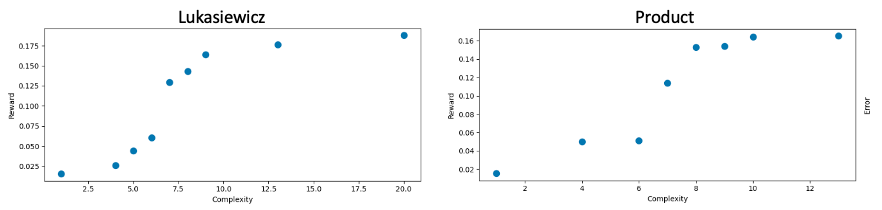}
    \caption{Pareto front showcasing the complexity versus reward for the Łukasiewicz implication versus the Product implication.}
    \label{fig:pareto_comparison}
\end{figure}

The simpler structure of the Product formula means fewer nested operations, making it more transparent how each feature affects the outcome. The use of basic fuzzy operations without excessive nesting helps in understanding the logical flow from input features to the fraud likelihood score. This can also be see in Figure \ref{fig:pareto_comparison}, where the complexity of formulas is generally lower for the Product implication compared to the Łukasiewicz implication.

Balancing the complexity and explainability of fuzzy logic expressions is crucial. While more complex formulas like the one using the Łukasiewicz implication can capture detailed patterns in data, they can become challenging to interpret. Simpler formulas, such as those using the Product implication, offer better transparency but might miss some intricate relationships. The Pareto front approach assists in selecting the optimal balance between these trade-offs, ensuring that the model remains both effective and explainable.

\section{Conclusion}
\label{sec:conclusion}
This paper explored the integration of fuzzy logic into DSR to enhance the explainability and performance of fraud detection models. In particular, using the obtained framework, we compared different fuzzy logic implications. Furthermore, we investigated the alternatives of training DSR on different reward functions to tailor to specific situations, particularly regarding highly imbalanced datasets. Moreover, we analysed the performance of unconstrained and constrained implementations of fuzzy logic to gather insights into manipulating fuzzy implications to alter performance and explainability.

\noindent\textbf{Comparative Analysis of Fuzzy Logic Implications.} 
It is noteworthy to mention that the Łukasiewicz implication consistently outperforms the Product, Gödel and combined implications in terms of raw performance, answering Question \ref{rq:sub1}. This can be concluded from both the single best expression results in Table \ref{table:bestresults} and average expression results in Table \ref{table:avgresults}.
However, when evaluating the implications and their complexity, we draw a different conclusion. The Łukasiewicz implications are generally harder to interpret compared to other implications. Although the performance of the Product implication is lower, both in the best expression and the average expression, the complexity associated with this implication is much lower. The average complexity for Łukasiewicz is 15.9, whereas the Product implication averages 13.9. Furthermore, the best performing expression of the Product implication has a complexity of 10, compared to a complexity of 20 for the Łukasiewicz implication. Therefore, we deem the Product implication the most effective in enhancing the model's ability to handle and explain the inherent complexity and uncertainty in fraud detection datasets while also improving explainability, answering Question \ref{rq:sub2}.

\noindent\textbf{Enhancing Model Explainability through Fuzzy Logic}
Integrating fuzzy logic into DSR provides compact formulas, however the obtained formulas truly require deeper look; and they are in general not very easy to interpret themselves.  However, the potential seems to exist for getting clearer insights into the model's decision-making process, thereby improving the explainability of the fraud detection model. While the expressions generated by this research are not inherently easy to interpret, they could offer possibly valuable insights when analyzed by experts. 

\noindent\textbf{Trade-off between Complexity and Performance}
Our findings reveal a trade-off between the complexity of fuzzy logic formulas and their performance. Unconstrained implementations tend to produce more complex formulas with better performance, while constrained implementations simplify the formulas at the cost of reduced performance. Utilizing Pareto front analysis and expert interpretation helps in balancing predictive accuracy with formula simplicity, ensuring that the models remain both effective and explainable, answering Question \ref{rq:sub4}. Experts should evaluate the expressions and choose a fuzzy logic formula according to the performance, explainability and suitability to different scenarios.

\noindent\textbf{Limitations and their Impact on Conclusions}
Despite the promising results, our study has limitations that must be acknowledged. The performance of fuzzy logic implementations was lower than SOTA methods, primarily due to the information loss when transforming continuous data into fuzzy sets based on percentiles. Future work should explore advanced methods like fuzzy clustering and alternative data transformations to retain more information.

Additionally, computational constraints limited the thoroughness of hyperparameter tuning and the complexity of the generated expressions. Addressing these constraints will be essential for enhancing the efficiency and scalability of the DSR framework.


Future research agenda is to focus on several key areas to enhance the current framework: including fuzzy clustering and data transformations, and investigating integrating fuzzy logic and related formalisms into newer symbolic architectures.
\bibliographystyle{ACM-Reference-Format}
\bibliography{references}

\end{document}